# Deep Domain Adaptation Based Video Smoke Detection using Synthetic Smoke Images

Gao Xu, Yongming Zhang, Qixing Zhang*, Gaohua Lin, Jinjun Wang

State Key Laboratory of Fire Science, University of Science and Technology of China, Hefei 230026, China

*Corresponding author: Qixing Zhang, Email: qixing@ustc.edu.cn, Tel: 86-551-63600770

## Abstract

In this paper, a deep domain adaptation based method for video smoke detection is proposed to extract a powerful feature representation of smoke. Due to the smoke image samples limited in scale and diversity for deep CNN training, we systematically produced adequate synthetic smoke images with a wide variation in the smoke shape, background and lighting conditions. Considering that the appearance gap (dataset bias) between synthetic and real smoke images degrades significantly the performance of the trained model on the test set composed fully of real images, we build deep architectures based on domain adaptation to confuse the distributions of features extracted from synthetic and real smoke images. This approach expands the domain-invariant feature space for smoke image samples. With their approximate feature distribution off non-smoke images, the recognition rate of the trained model is improved significantly compared to the model trained directly on mixed dataset of synthetic and real images. Experimentally, several deep architectures with different design choices are applied to the smoke detector. The ultimate framework can get a satisfactory result on the test set. We believe that our approach is a start in the direction of utilizing deep neural networks enhanced with synthetic smoke images for video smoke detection.

## 1. Introduction

Video smoke detection, as a promising fire detection method especially in open or large spaces and outdoor environments, has been researched more than ten years [1, 2]. The traditional video smoke detection methods apply the pattern recognition technology such as feature extraction and classifier design, whose main focus is on the combination of static and dynamic characteristics for smoke recognition. The typical characteristics contains color and motion probability [3], histogram sequence of LBP for smoke [4], dynamic texture using surfacelet transform and HMT model [5], smoke temporal trajectory [6], local extremal region segmentation [7], and motion orientation model [8], etc. Compared to these methods based on traditional feature extractors designed manually, some researchers proposed to apply the deep convolutional neural network (CNN) to video smoke detection [9, 10], which can learn a deeper characteristic model with higher generalization capability. As far as we know, the datasets these researchers used only contains thousands of smoke images, which is unfavorable for deep learning.

Due to the smoke images samples limited in scale and diversity for CNN training, we use synthetic smoke images to extend the training set. Similar work is described in [11], which focused on synthesizing wildfire smoke images by image overlay processing with frame of virtual environment. Compared with them, we pursue the diversity of synthetic smoke image rendered with background. For our task, the practical benefit of synthetic smoke images is to increase recognition power of the model trained on them, rather than their visual effects.

As there is certain appearance gap between synthetic and real smoke images, the difference of their statistical distributions can degrade the performance of the trained model on test set composed of real images. In order to tackle this problem and make efficient use of synthetic smoke images, we apply the domain adaptation (DA) method to build the deep architecture, which acts on confusing the distributions of features extracted from synthetic and real smoke images. This method expands the domain-invariant feature space of smoke images off non-smoke images. To our knowledge, there is no study to apply synthetic smoke images with relevant deep architectures based on domain adaptation to the video smoke detection.

The main process of our work is as follows. Firstly, we build a synthesis pipeline and set various conditions for smoke simulation and rendering randomly to produce synthetic smoke images of high diversity. Secondly, the whole dataset is divided into source dataset (synthetic smoke and non-smoke) and target dataset (real smoke and non-smoke). Multi-label deep architecture is built based on domain adaptation, which is designed to extract the domain-invariant features from synthetic and real smoke images. Finally, experiments are conducted to evaluate the performance of architectures with different design choices.

## 2. Related Work

### 2.1. Deep learning and synthetic image

Deep learning methods have achieved a state-of-the-art performance on the computer vision. Compared to the traditional methods to extract shallow features manually, the deep architecture obtains the more essential features independently. The deep learning technology has made breakthroughs in many field. For example, Tang proposed DeepID3 [12], whose face verification precision achieves to 99.53% on LFW (Labeled Faces in the wild). Typically, the team of Microsoft Research developed a residual network [13] and achieves 3.57% error rate on the ILSVRC 2015 classification task, which is lower than the 5.1% error rate of the human eyes. This indicates that the deep learning has surpassed the human-level performance in the task of image classification.

The current deep learning methods is trained and tested on the rich dataset, such as ImageNet [14] and LFW. In this paper, we use adequate synthetic smoke images of high diversity to extend the training set with real smoke images captured from experiments. With the development of CG technology, synthetic image has been widely used in machine learning [15-18]. For instance, due to the lack of labeled images for 3D recognition, Yu [19] synthesized the dataset called PASCAL 3D+ for object detection and viewpoint estimation. The images in PASCAL 3D+ exhibit much more variability compared to the existing 3D datasets. A related work in [20], which designed a scalable and overfit-resistant image synthesis pipeline to expand PASCAL 3D+ dataset. But their object is 3D rigid models and they train CNN directly on the generated data. Experimentally, the model trained on these synthetic images can't perform well on real images, especially for smoke images (see Table 1).

**2.2. Domain adaptation**

When the training data have different statistical distribution from the test data which the trained model is applied on, the performance of classifier will be degraded. In our case, the model trained mainly on synthetic images performs undesirable on real images, due to the dataset bias between synthetic and real image dataset (domain). Domain adaptation [21] tackles this problem by reducing the domain variation of features extracted in the related dataset. To ensure that model trained on synthetic images has a good performance on the real images, Baochen [22] presented a supervised adaptation approach based on de-correlated features, and their result showed that non-photorealistic synthetic images works as well as more realistic synthetic images when training detectors. Saenko [23] proposed to learn a linear transformation from source domain to target domain to adapt features in classification algorithm. Eric [24] set an adaptation layer and an additional domain confusion loss function layer in the CNN architecture to learn a feature representation both discriminative and domain invariant. Besides in [25], they designed a matching loss function to transfer class-correlation learned on the source samples to the target samples. Wen [26] transformed the features of each domain data into a common subspace, and compute the mapping matrix by minimizing the structural risk function of SVM.

For video smoke detection, smoke recognition is a binary classification task. We only synthesize smoke images while non-smoke images are all real images. In this case, we show our approach based on domain adaptation can obtain well result when tested on the multifarious real smoke and non-smoke images.

## 3. Smoke image dataset for CNN training

Available smoke images for training set are obtained generally from Internet and experiments, which are limited in scale and diversity for training deep CNN model. We build a synthesis pipeline (see Fig. 1) to produce synthetic smoke images of high diversity. The production process is automated in Blender-python.

Compared to the pipeline for rendering the 3D rigid models in [20], visual simulation of smoke is more complex as the representation of synthetic smoke image is determined by numerical simulation and media rendering for smoke. Especially, due to the fuzzy transparency of early smoke, it is necessary to render smoke with background image instead of composition of them.

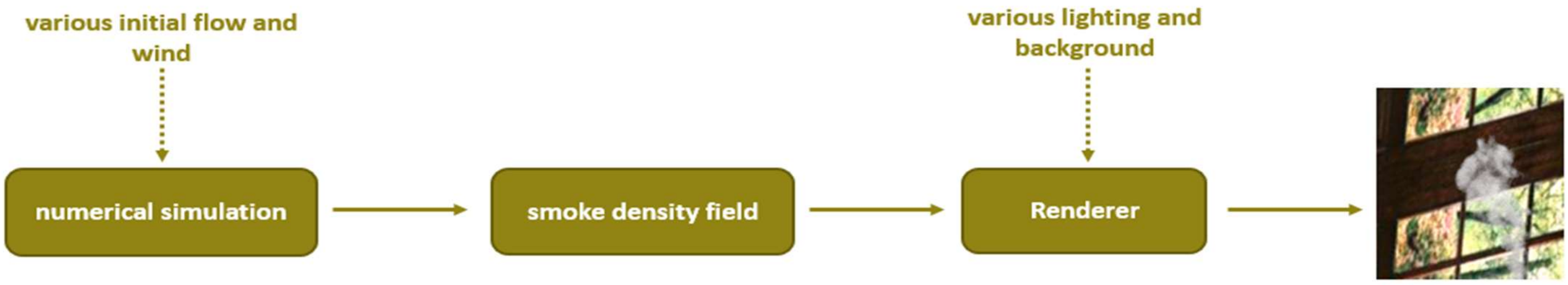

Fig. 1. Synthesis pipeline overview. To increase the diversity, the initial flow, wind, lighting and background are set randomly.

Another work is to extract the smoke region, as our recognition object is separated rectangular region of smoke. For synthetic smoke image, it is easy to extract the smoke region because the location of smoke domain is known. And for real smoke image, we manually selected the smoke region.

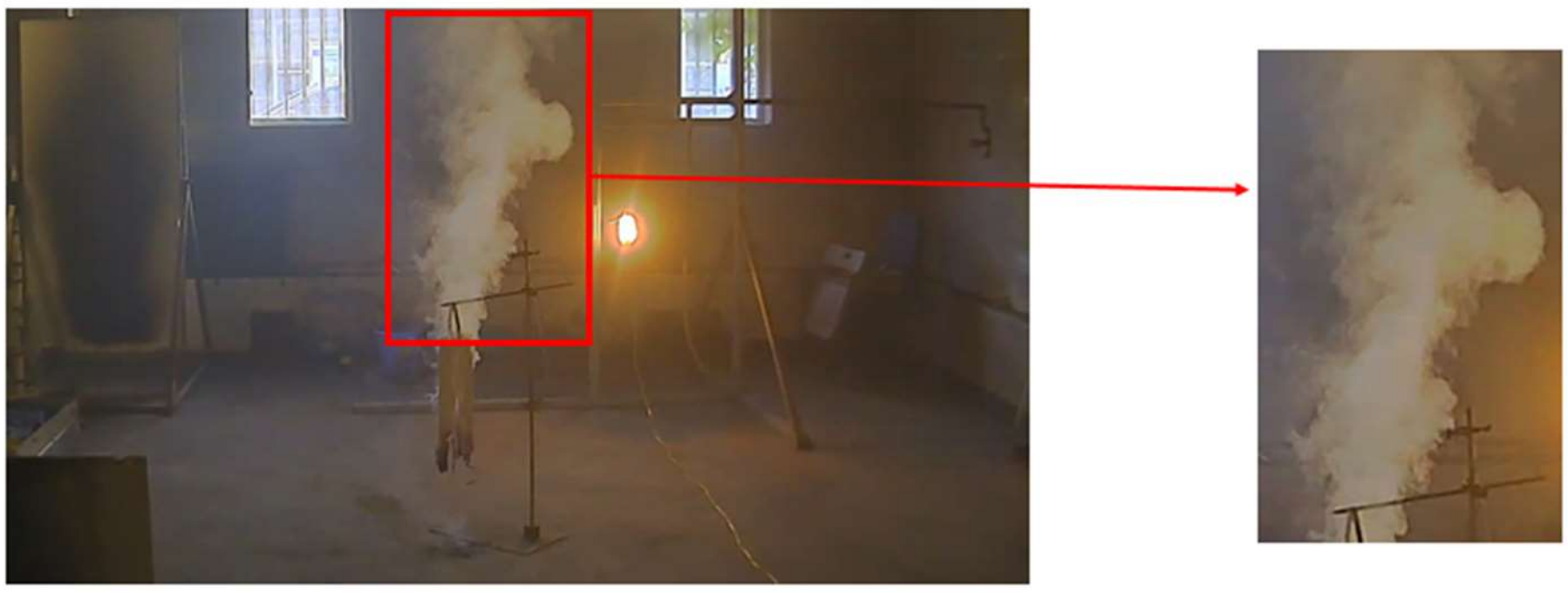

Fig. 2. Extraction for separated rectangular region of smoke.

On the one hand, smoke videos were captured under conditions indoor and outdoor. 5K real smoke images were selected every 5 or 25 frames from the videos. One the other hand, 30K synthetic smoke images of high diversity were produced. Instead of the method to split smoke blocks, we cropped out the entire rectangular region of smoke. Meanwhile, the same number of non-smoke images were collected. In addition, a test set including 1000 images is created to evaluate the performance of the trained model. As shown in Fig. 3, in order to measure the robustness of the methods, the test smoke images are quietly different from the real smoke images in our training set and the test non-smoke images have a strong interference to the smoke recognition.

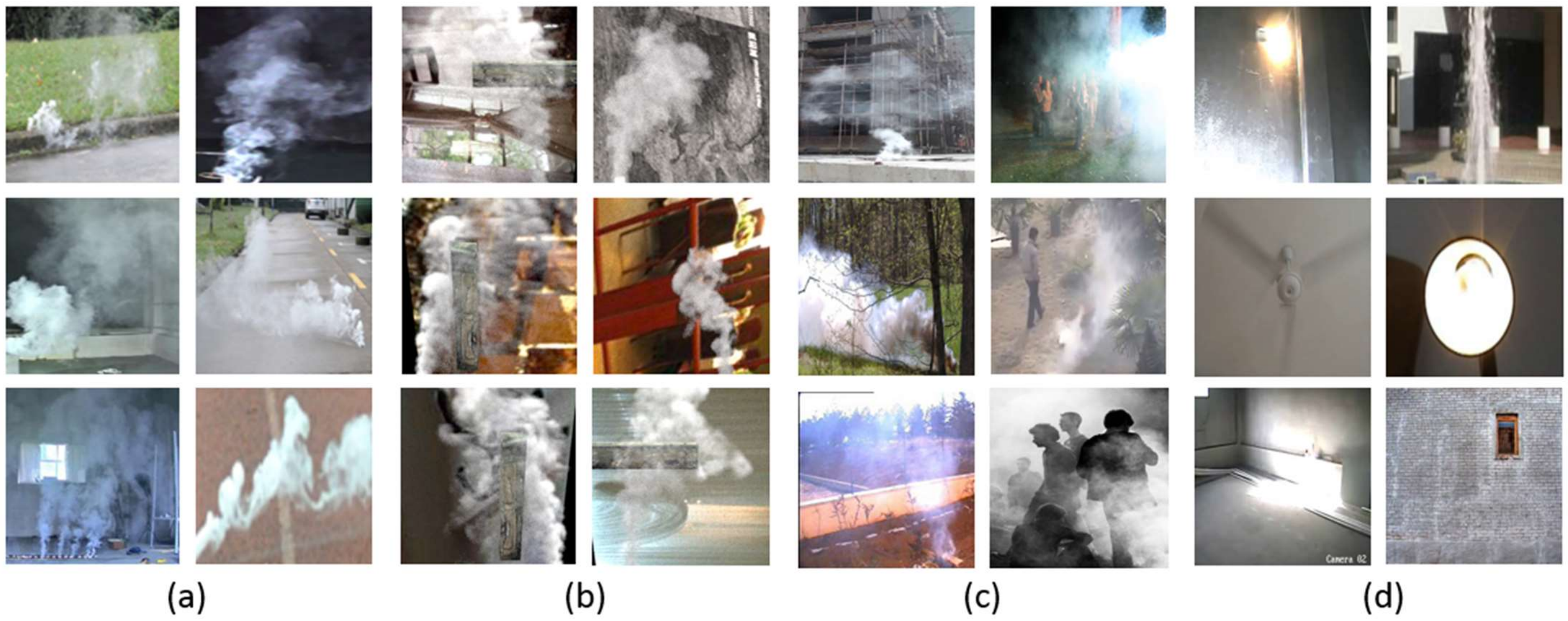

Fig. 3. Image samples: (a) real smoke images; (b) synthetic smoke images; (c) test smoke images; (d) test non-smoke images.

## 4. Network and layer function

As there is a certain appearance gap between synthetic and real smoke images, essentially the difference between their feature distributions degrades significantly the performance of trained model when tested on real images in practical application. In order to tackle this problem and learn a strong feature representation from the synthetic smoke images, our deep architecture is constructed based on domain adaptation, which is the standard approach to alleviate dataset bias caused by a difference in the statistical distributions between training and test data. In this way, not only the deep features extracted are discriminative between smoke and non-smoke images, but also invariant with respect to the shift between the synthetic and real smoke images.

As shown in Fig. 4, the whole dataset is divided into source dataset and target dataset. In general, these two datasets for domain adaptation respectively consist of labeled data entirely from one domain (e.g. synthetic images) and unlabeled data entirely from another domain (e.g. real images). But in our work, all the samples are labeled and the non-smoke image samples in these two datasets are all real images. In detail, the source dataset contains synthetic smoke images and real non-smoke images, and the target dataset contains real smoke images and real non-smoke images. In this case, we denote with multi-label $(y_i^s, y_i^d)$ for each sample $x_i$ of these two datasets, in which $y_i^s$ indicates whether $x_i$ is a smoke image or not, $y_i^d$ indicates whether $x_i$ is a real or synthetic image.

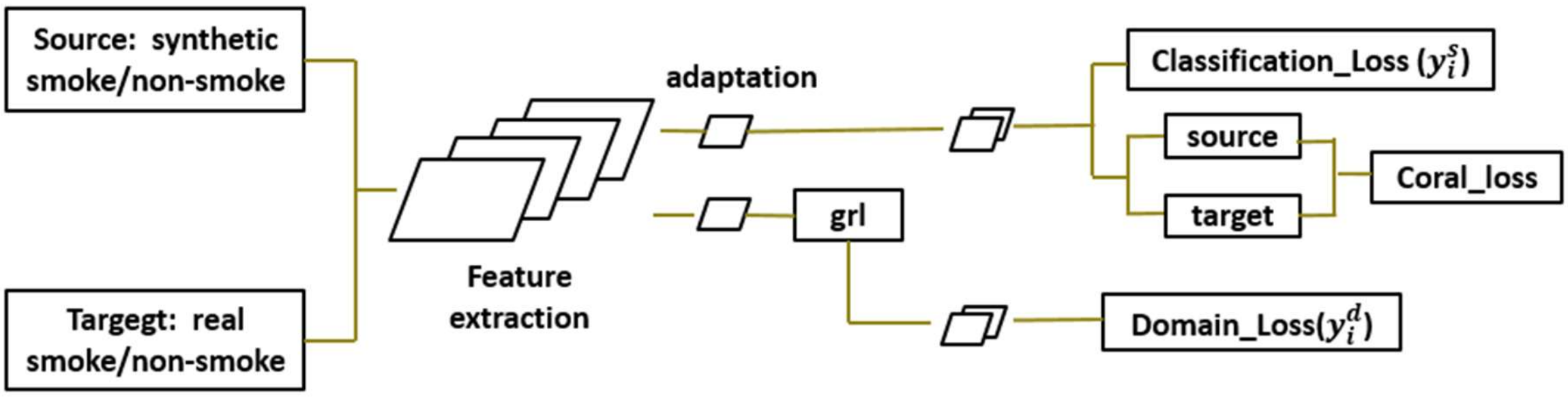

Fig. 4. This CNN architecture contains shared feature extraction layers, feature adaptation layers, and three loss function layers.

Different from setting the classifier only for labeled source data in the general domain adaptation method, the classifier (whether it is smoke image) is set for data from the two datasets.

We use the feature extraction layers of AlexNet and the softmax loss is defined as the classification loss of the label $y_i^s$,

$$L_s = -\frac{1}{N}\sum_{1}^{N}\log[soft\max(\alpha_i)] \tag{5}$$

where $\alpha_i$ is the predicted probability of the input sample $x_i$ on the label $y_i^s$, and N is the size of a batch.

Besides, we take the hinge loss as the domain loss of the label $y_i^d$,

$$L_d = \frac{1}{N}\sum_{1}^{N}(\max(0,1-\sigma(l_i = k)t_{ik}))^p \tag{6}$$

where $l_i$ is the predict label, $\sigma(l_i = k) = \begin{cases} 1, & k = y_i^d \\ -1, & k \neq y_i^d \end{cases}$, and the $t_{ik}$ is the predicted probability of $x_i$ on the label k.

At the training time, our architecture needs to achieve two objectives: minimize the classification error $L_s$ to obtain discriminative feature representation and maximize the domain loss $L_d$ to confuse (close) the distributions between synthetic and real smoke images through gradient reversal layer(GRL)[27]. The GRL plays the all-pass function during forward and multiplies the gradient by $\varphi$ during backpropagation.

In [24], an adaptation layer is added to follow the feature extraction layers to directly adapt the representation. This layer is used to prevent overfitting to the particular nuances of the distribution of synthetic images in source dataset.

The adaptation layer implements transformation to project the source and target distributions into a lower dimension space better for the tasks of classification and domain confusion:

$$f_{adapt} = Af_{represent} + x \tag{7}$$

This transformation is easily achieved in CNN to reform the fully-connected layer.

Experiments show that it is difficult to obtain a satisfactory equilibrium between the two objectives (see Table 2). This supervised training only aligns roughly the distributions of feature extracted from synthetic and real smoke images in statistical distribution, namely it almost aligns their means but not the every local distribution. Therefore, the correlation of their features cannot be guaranteed. It cannot extract a feature representation whose distribution achieve our goal. To confuse fully the feature distributions between them, we added a CORAL [28] loss layer to align the second-order statics of the source and target feature distributions for correlation alignment to make them closer,

$$L_{coral} = \frac{1}{4d^2} \left\| C_S - C_T \right\|_F^2 \tag{8}$$

where $C_S$ and $C_T$ are covariance matrices of source and target feature representations of $d$-dimension.

The joint loss function of our architecture is as follow,

$$L = \alpha_{label} * L_s + \beta_{domain} * \varphi * L_d + \gamma_{coral} L_{coral} \tag{9}$$

where the loss weight $\alpha_{label}$, $\beta_{domain}$ and $\gamma_{coral}$ determine how strongly the three loss functions influence the optimization.

## 5. Experiments

In this section, the effectiveness of synthetic smoke images to training is confirmed and relevant evaluation of the domain adaptation based deep architecture are performed. At first, we represent the predicted results on test set of models trained on different datasets. Then the comparison experiments of the domain adaptation based deep architecture with different design choices are conducted to evaluate their function on recognition rate. In addition, the factors that influence the recognition power of model in the training process are analyzed and experiments give the investigation on the design choice of architecture which dominate the deep CNN model to best performance.

## 5.1. Evaluation

It is essential to clarify the effect of synthetic smoke images to the detectors. We used the network structure of AlexNet and train the model on different datasets. Note, the output of classification is the probability of each sample on the two categories, and we take the category with larger probability as the predicted label. In order to qualitatively and quantitatively evaluate the performance of model at the whole test set, the correct detection(CD), the error detection(ED) and missed detection(MD) are measured in Table 1. The three values are defined as:

$$\mathrm{CD} = \frac{number\ of\ true\ positive\ and\ true\ negative}{1000} \times 100\%$$

$$\mathrm{ED} = \frac{number\ of\ false\ positive}{number\ of\ predict\ label = smoke} \times 100\%$$

$$\mathrm{MD} = \frac{number\ of\ false\ negative}{500} \times 100\%$$

Note: there is 500 smoke images and 500 non-smoke images in the test set.

Table 1. The performance of model of AlexNet trained on different datasets.

| Training set(contains non-smoke images) | CD | ED | MD |
|---|---|---|---|
| Real smoke images | 0.6690 | 0.0526 | 0.6420 |
| Synthetic smoke images | 0.5700 | 0.2160 | 0.8060 |
| Mixed dataset of real and synthetic smoke images | 0.7380 | 0.0162 | 0.5160 |

Due to the limitation of real smoke images in scale and diversity, as there is only 5k real smoke images, the model trained on them miss nearly 64% of the test smoke images which are quite different from the training smoke images. Because of this limited training, the model does not identify these “strange test images” as smoke and the value of the ED is low.

Meanwhile, the performance of the model trained on the synthetic smoke images (all of the 30K) is more terrible. Although synthetic smoke images can provide a richer diversity, after all the synthetic smoke images are certain different on appearance from real images. Therefore, we train the model on the mixed dataset which consists of almost the same number of real and synthetic smoke images. The results show that the predictive accuracy is slightly improved. Judging from the above results, the model trained directly on the mixed dataset benefits little from the synthetic smoke images.

The values of the CD and MD are not enough to meet the actual requirements. It is in this case that we use the domain adaptation based deep architecture to tackle this problem. The whole dataset is divided into the source and target datasets. The target dataset contains 5k real smoke images, and the source dataset contains synthetic smoke images from 5k to 30k. We will analyze the impact of the composition of dataset on the predicted results in the following section. Several deep architectures based on domain adaptation are trained on them. The predicted results of architectures with different design choices are reported in Table 2.

Table 2. The performance of different deep architectures based on domain adaptation.

| Architecture<br>(the layers added to feature extraction layers) | CD | ED | MD |
|---|---|---|---|
| Ours with GRL | 0.8170 | 0.1768 | 0.1920 |
| Ours with GRL + adaptation layer for $L_d$ | 0.8080 | 0.2079 | 0.1640 |
| Ours with GRL + adaptation layer for $L_s$ and $L_d$ | 0.8520 | 0.1633 | 0.1240 |
| Ours with GRL + CORAL with adaptation layer | 0.9470 | 0.0447 | 0.0620 |

It can be seen that the predicted accuracy of the models of these deep architectures based on domain adaptation are improved significantly than that of the general architecture trained on the mixed dataset. These architectures all use the gradient reversal layer (GRL) to connect to domain loss layer, for confusing the features of synthetic and real smoke images. By comparison, the adaptation layer is actually useful as it avoids the overfitting in synthetic smoke images, when added on the bottom of the layers connected to the $L_s$ $and$ $L_d$ loss layers.

In our experiment, the last architecture achieved the best performance. As we describe in section 4, the CORAL loss layer make the distributions of the two datasets closer, especially for the smoke image samples in the two datasets as the non-smoke image samples are basically the same. Of course, better performance may be obtained to adjust the architecture and training.

In order to represent the distributions of features extracted by adaptation architecture more intuitive, t-SNE [29] visualizations of feature distributions are showed in Fig. 5. These point cloud show the effect of network on the distribution of deep feature representations of synthetic smoke、real smoke and non-smoke images. Obviously, as shown in Fig. 5(i), the features of synthetic smoke images are easily confused with non-smoke images rather than real smoke images. By comparison, the feature distributions of real and synthetic images in Fig. 5(ii) are aligned and confused fully, separated from non-smoke images. This case has a fine performance as it expand the domain-invariant feature space for smoke images as expected.

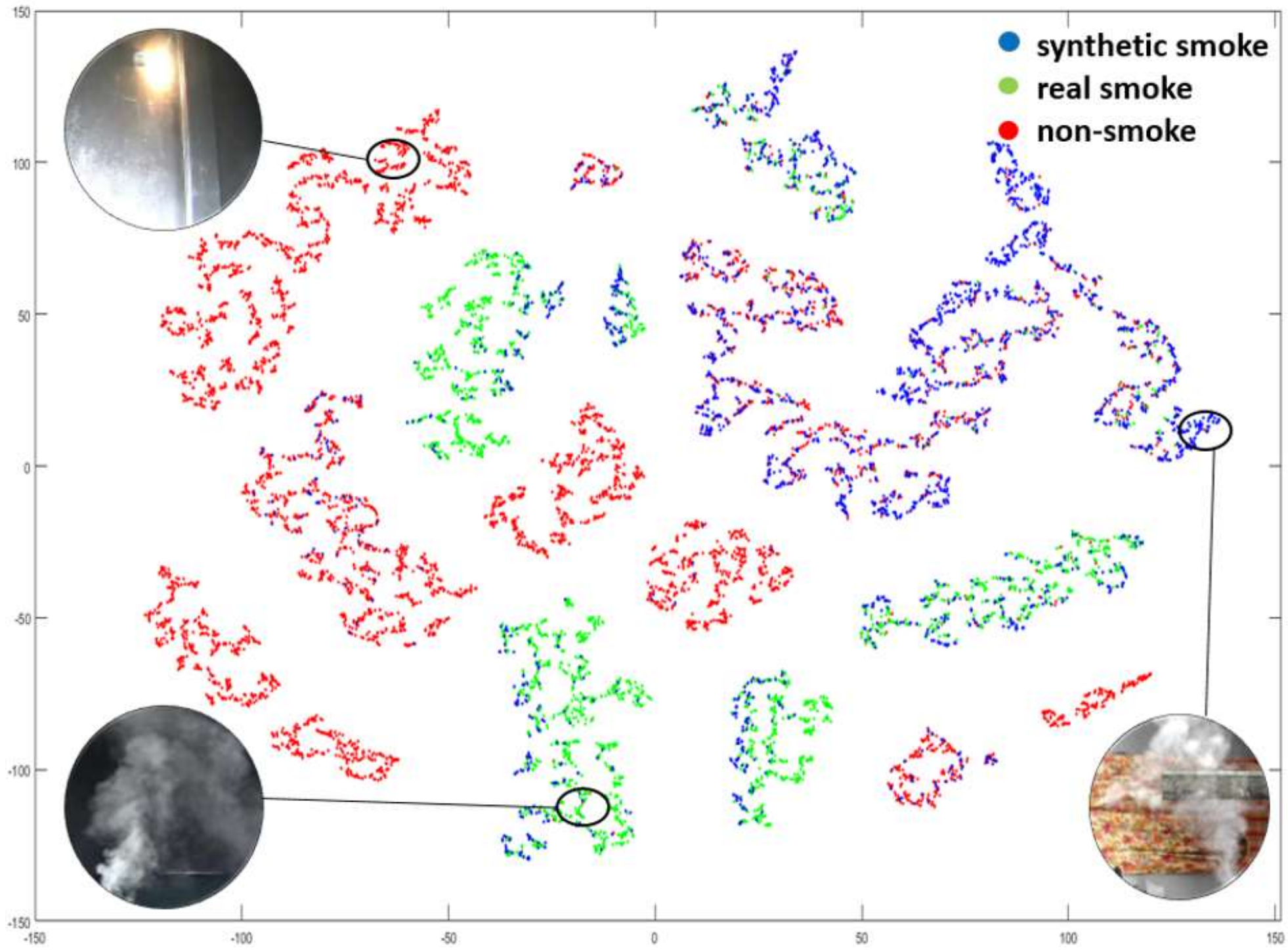

(i) Results obtained with the model of Alexnet trained on mixed dataset

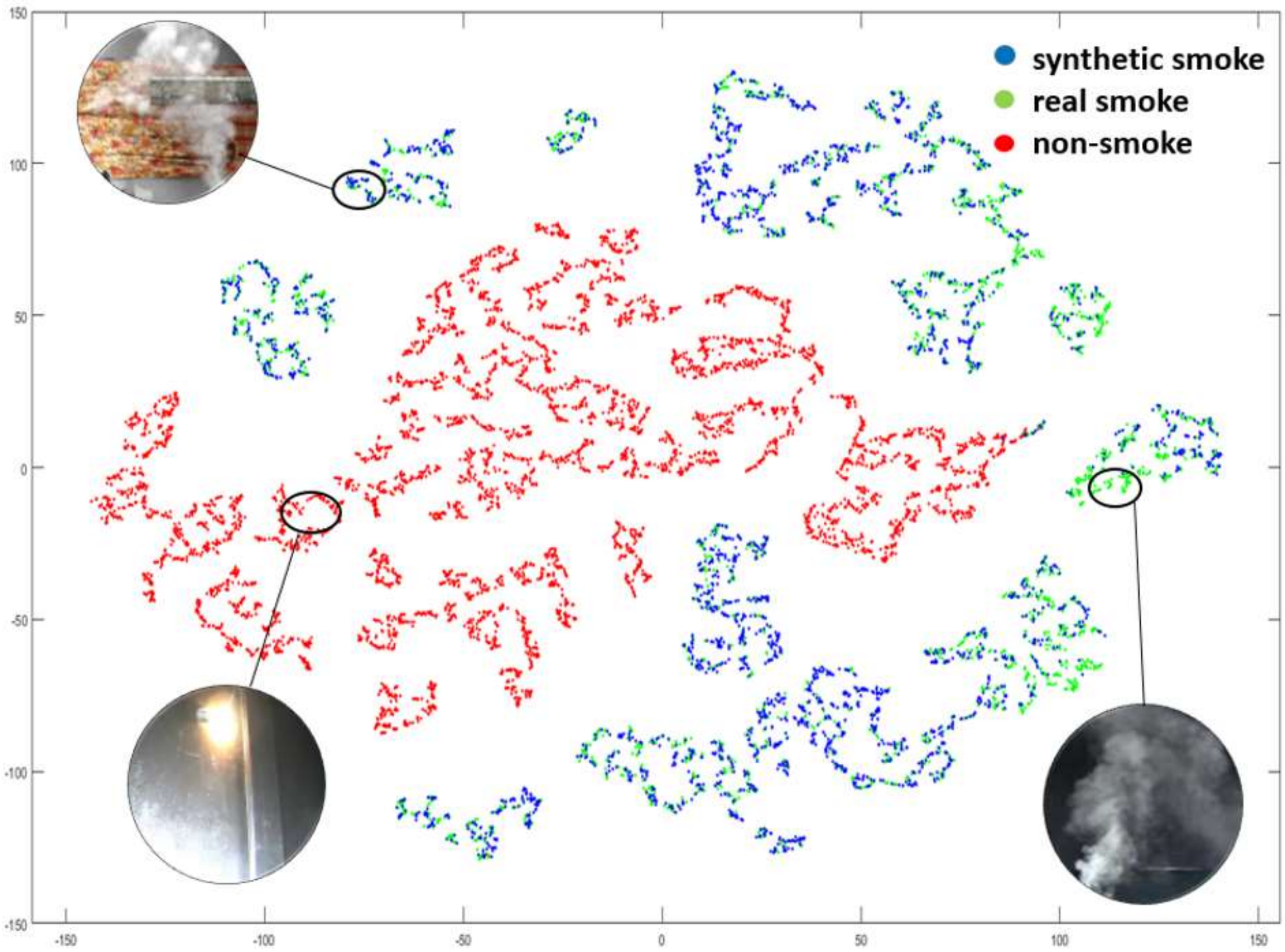

(ii) Results obtained with the model of adaptation architecture trained on the source and target datasets

Fig. 5. The visualizations for feature distributions of synthetic smoke (blue), real smoke (green) and non-smoke (red) image samples.

## 5.2. Investigation

Some researchers [16] have investigated the effect of the proportion of the synthetic image to the real image in the mixed dataset on the recognition results. Similar effect maybe exist in the domain adaptation based deep architecture. In this test, the scale proportion of the non-smoke to smoke images is 1. We increase the scale of the target dataset to 20k by data augmentation. Then, the different scale of source dataset is set from 20k to 60k. As shown in Fig. 6, the predicted results indicated that the scale proportion of the source dataset to target dataset makes little impact on the predictive accuracy. But the scale of source dataset is too larger than target dataset, namely, the synthetic smoke images is much more than real smoke images, the value of ED will increase. Meanwhile, Fig. 7 shows that the scale proportion of the non-smoke images to smoke images in the two datasets has a certain effect on the ED and MD with a shift between them.

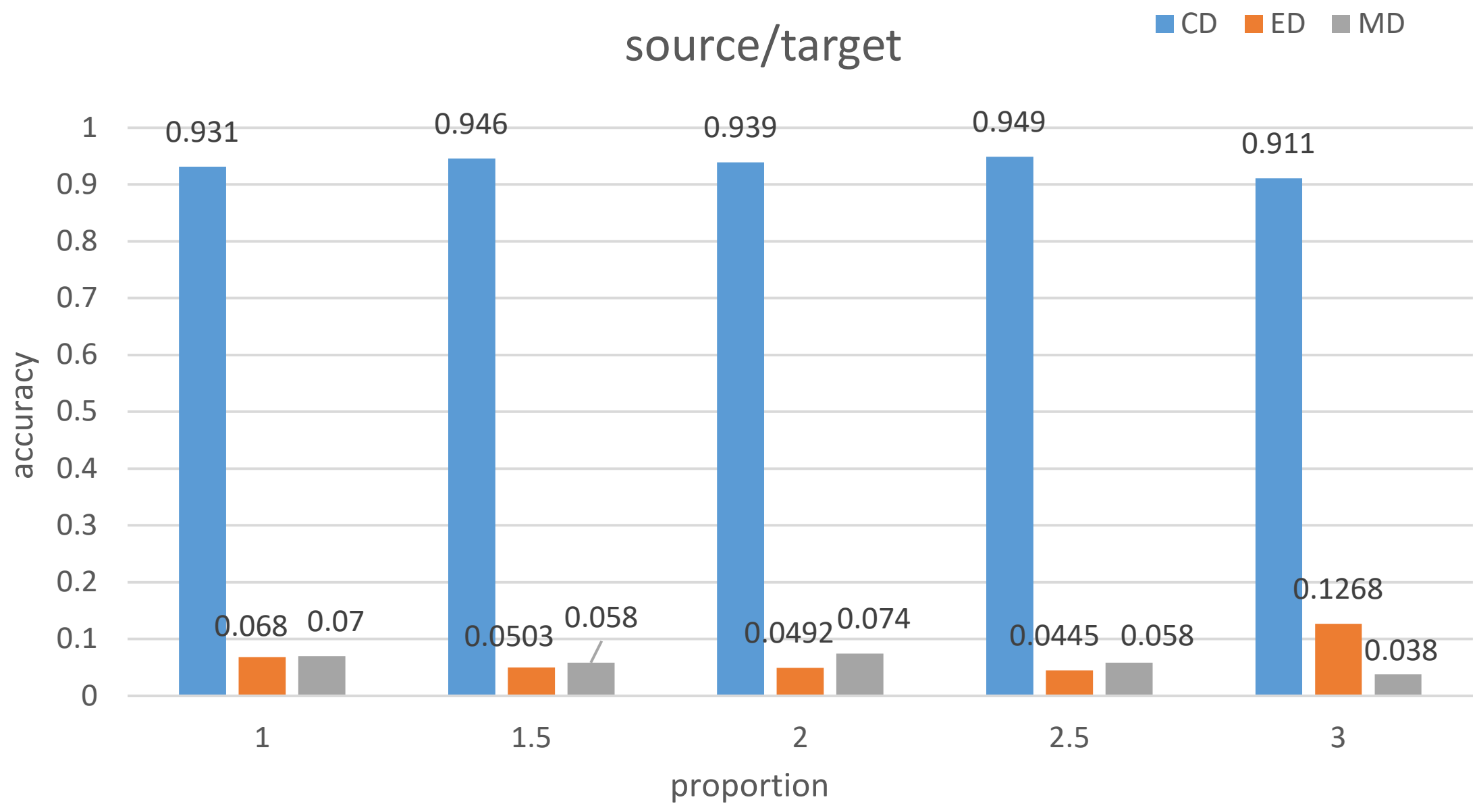

Fig. 6. The test accuracy for different scale proportion of the source dataset to target dataset.

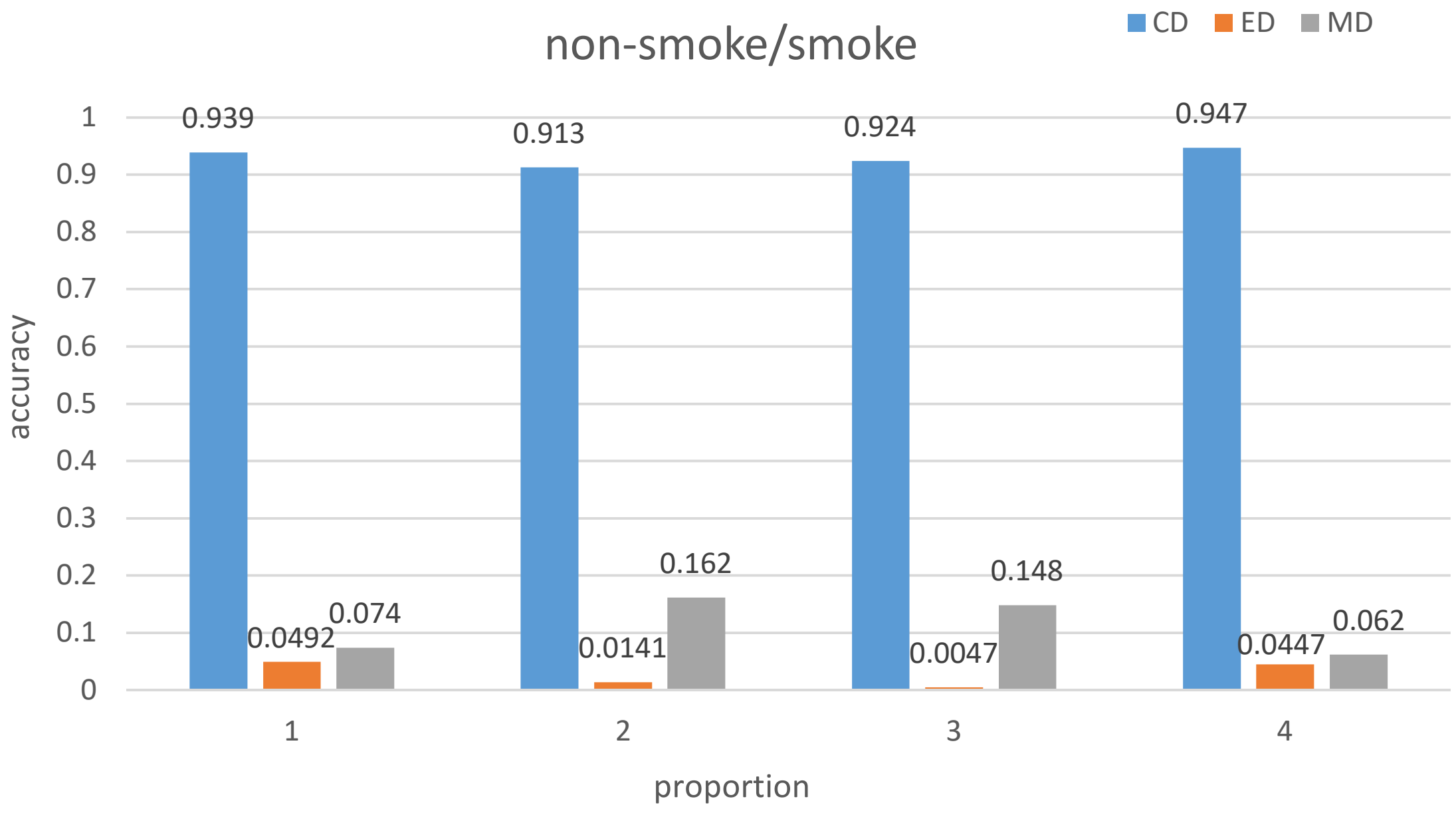

Fig. 7. The test accuracy for different scale proportion of the non-smoke to smoke image samples.

In the training process, the loss $L_s$, $L_d$ and $L_{coral}$ are fed back simultaneously to the weight update of shared feature extraction layers during the backpropagation. Accordingly, the loss weight $\alpha_{label}$, $\beta_{domain}$ and $\gamma_{coral}$ determine their influence. In this test, we focused on the discrimination between smoke and non-smoke images, and confusion between the synthetic and real smoke images. In order to facilitate comparison between the two effects, $\alpha_{label}+\beta_{domain}$ = 1 is set. Fig. 8 shows the performance of the model trained with different value of loss weight $\beta_{domain}$. The results show that the loss of classification should be still dominant in the training.

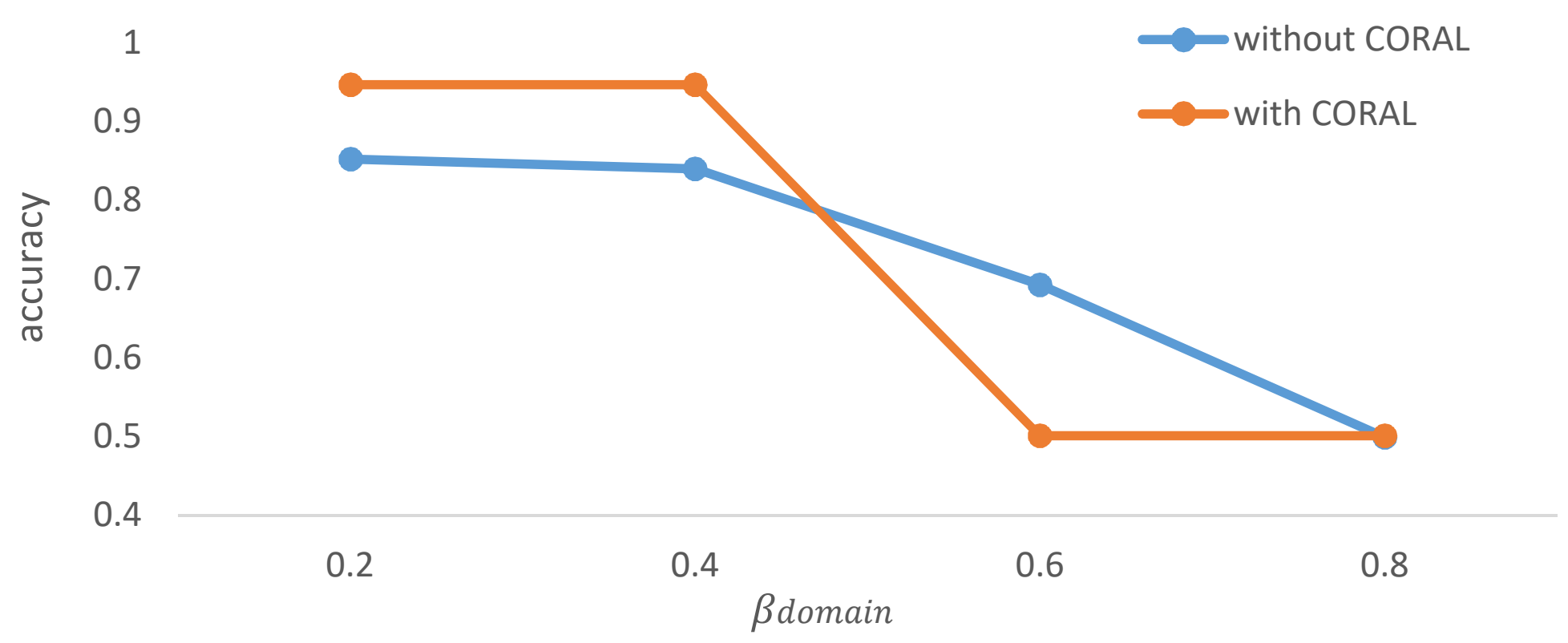

Fig. 8. The test accuracy for different value of loss weight $\beta_{domain}$. The blue curve represents the performance of the model trained without correlation alignment, and the orange curve represents the performance of the model trained with correlation alignment ($\gamma_{coral} = 0.2$).

## 6. Conclusion

We propose a deep domain adaptation based approach for video smoke detection to extract a power feature representation of smoke. Due to the smoke image samples limited in scale and diversity for training deep model, we systematically produced adequate synthetic smoke images rich in variation. In order to prevent the degradation in performance of trained model caused by the appearance gap between synthetic and real smoke images in training set, we apply the domain adaptation based deep architecture to the classification task. Experiments confirmed the effectiveness of synthetic smoke images to training and investigate the effects of different composition of dataset and design choices of the deep architecture on the predicted results. The ultimate framework can get a satisfactory result on the test set. As the appearance of synthetic smoke images directly affects the performance of trained model, in the future work, we will investigate the effect of variations caused by synthesis on recognition power of model and our method maybe benefit from it.

## Acknowledgements

This work was supported by the National Key Research and Development Plan under Grant No. 2016YFC0800100, the National Natural Science Foundation of China under Grant No. 41675024, and the Fundamental Research Funds for the Central Universities under Grant No. WK2320000033 and No. WK6030000029. The authors gratefully acknowledge all of these supports.